\documentclass{sig-alternate}
\newfont{\mycrnotice}{ptmr8t at 7pt}
\newfont{\myconfname}{ptmri8t at 7pt}
\let\crnotice\mycrnotice%
\let\confname\myconfname%

\usepackage{times}
\usepackage{epsfig}
\usepackage{graphicx}
\usepackage{amsmath}
\usepackage{amssymb}
\usepackage{algorithm}
\usepackage{algorithmic}
\usepackage{subfigure}
\usepackage{url}
\usepackage{multirow}
\usepackage{flushend}
\usepackage{booktabs}
\newcommand{\tabincell}[2]{\begin{tabular}{@{}#1@{}}#2\end{tabular}}

\begin{document}
%

\title{EventNet: A Large Scale Structured Concept Library for Complex Event Detection in Video}

\numberofauthors{1}
\author{\alignauthor Guangnan Ye{\large\thanks{Equal contribution.}}, Yitong Li{\large$^\ast$}, Hongliang Xu, Dong Liu, Shih-Fu Chang \\
\vspace{1mm}
\affaddr{
        Department of Electrical Engineering, Columbia University, New York, NY 10027, USA}\\
\vspace{1mm}
\email{\{gy2179, yl3029, hx2168, dl2713, sc250\}@columbia.edu}\\
\vspace{1mm}
}

\maketitle
\begin{abstract}
Event-specific concepts are the semantic concepts specifically designed for the events of interest, which can be used as a mid-level representation of complex events in videos. Existing methods only focus on defining event-specific concepts for a small number of pre-defined events, but cannot handle novel unseen events. This motivates us to build a large scale event-specific concept library that covers as many real-world events and their concepts as possible. Specifically, we choose WikiHow, an online forum containing a large number of how-to articles on human daily life events. We perform a coarse-to-fine event discovery process and discover $500$ events from WikiHow articles. Then we use each event name as query to search YouTube and discover event-specific concepts from the tags of returned videos. After an automatic filter process, we end up with  $95,321$ videos and $4,490$ concepts. We train a \emph{Convolutional Neural Network} (CNN) model on the $95,321$ videos over the $500$ events, and use the model to extract deep learning feature from video content. With the learned deep learning feature, we train $4,490$ binary SVM classifiers as the event-specific concept library. The concepts and events are further organized in a hierarchical structure defined by WikiHow, and the resultant concept library is called \emph{EventNet}. Finally, the EventNet concept library is used to generate concept based representation of event videos. To the best of our knowledge, EventNet represents the first video event ontology that organizes events and their concepts into a semantic structure. It offers great potential for event retrieval and browsing. Extensive experiments over the zero-shot event retrieval task when no training samples are available show that the proposed EventNet concept library consistently and significantly outperforms the state-of-the-art (such as the $20K$ ImageNet concepts trained with CNN) by a large margin up to $207\%$. We will also show that EventNet structure can help users find relevant concepts for novel event queries that cannot be well handled by conventional text based semantic analysis alone.
The unique two-step approach of first applying event detection models followed by detection of event-specific concepts also provides great potential to improve the efficiency and accuracy of Event Recounting since only a very small number of event-specific concept classifiers need to be fired after event detection.
\end{abstract}


\section{Introduction}
The prevalence of video capture devices and growing practice of video sharing in social media have resulted in enormous
explosion of user-generated videos on the Internet. Such video data records daily events related to various aspects of human life. Generally, these events can be categorized into two categories: procedural event and social event. The former includes procedural videos such as ``make a cake'', ``groom a dog'' and ``change a tire'' while the latter includes social activities such as ``birthday party'' and ``flash mob gathering''.

\begin{figure}[t]
\centering
{\includegraphics[width=1\linewidth]{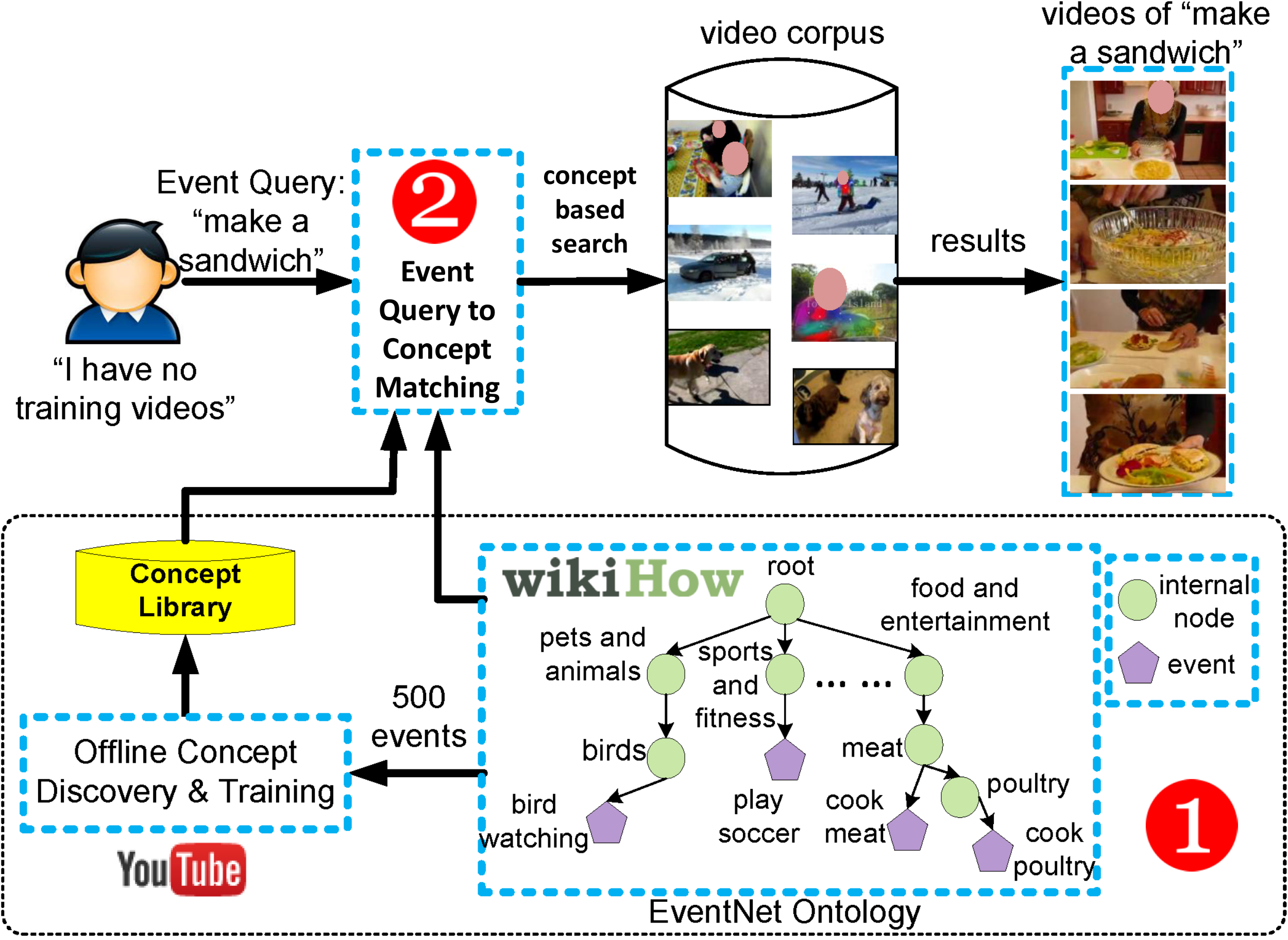}}
\caption{Concept based event retrieval by the proposed large scale structured concept library \emph{EventNet}. This paper presents two unique contributions: (1) A large scale structural event ontology. (2) Effective event-to-concept mapping via the ontology. }
\label{Fig:Framework}
\end{figure}

Existing video search systems are mostly based on textual descriptions of the videos such as tag keywords, metadata and surrounding text. Recent research in computer vision and multimedia attempts to automatically recognize events in videos, and index event videos based on the learned semantics. A notable example of such effort is TREC Video Retrieval Evaluation (TRECVID) Multimedia Event Detection (MED)~\cite{MED:NIST}.  Different from the action recognition task which focuses on primitive human activities such as ``walk'' and ``jump''~\cite{Laptev:CVPR08,Reddy:MVAP12}, event detection deals with complicated high-level human activities that interact with objects in a certain scene. The complexity of such event is further increased by the occlusion, motion, and scene transition seen in user generated videos. For example, a video of ``land a fish'' may contain many semantic concepts such as objects ``boat'' and ``fish'', actions ``land'' and ``grab'' as well as scenes ``pond'' and ``shore''. Most of the early works on event detection focus on building event detection models from low-level features while ignoring the rich semantics in event videos~\cite{Duan:CVPR10,Lai:ECCV14,Xu:CVPR14}. This makes analysis and interpretation of high-level events difficult, especially when there are only few or zero training videos at hand.

This deficiency has stimulated a new research direction that tries to represent event videos in a high-dimensional semantic space, using each dimension of the space to indicate the likelihood of the occurrence of a concept in the video~\cite{Liu:WACV13,Wu:CVPR14}. A key research problem is how to generate a suitable concept lexicon for events. Existing works can be categorized into two groups. The first is event independent concept lexicon, which directly borrows the object, scene and action concepts from existing works and applies them to video events~\cite{Wu:CVPR14}. However, since these concepts are not specifically optimized for events of interest, they are often insufficient in capturing the rich semantics contained in the events. The second is event driven concept lexicon, which uses the pre-known event definition as query keywords to search external resources such as Flickr and YouTube, and then discover concepts that are specifically relevant to the target events~\cite{Chen:ICMR14}. Although this approach achieves good performance, it can only deal with a small number of target events whose definitions are known beforehand. When a novel unseen event emerges, it is no longer applicable due to the lack of relevant concepts for the unseen event.

To address this problem, we propose to construct a large scale event-driven concept library that covers as many real world events and concepts as possible. Figure~\ref{Fig:Framework} illustrates the overall framework of the proposed method, in which we highlight two main challenges addressed in this paper. The first is how to define events and their relevant concepts in order to construct a comprehensive concept library. To achieve this goal, we resort to the external knowledge base called WikiHow~\cite{WikiHow}, a collaborative forum that aims to build the world's largest manual for human daily life events. We define $500$ events from the articles of WikiHow website. For each event, we use its name as query keywords to perform text based video search on YouTube, and apply our automatic concept discovery method to discover event-specific concepts from the tags of the returned videos. Then we crawl videos associated with each discovered concept as the training videos to learn deep learning video features using \emph{Convolutional Neutral Network} (CNN) as well as event-specific concept models. This leads to an event-specific concept library composed of $4,490$ concept models trained over $95,321$ YouTube videos. We further organize all events and their associated event-specific concepts into a hierarchical structure defined by WikiHow, and call the resultant concept library \emph{EventNet}.

The second challenge is how to find semantically relevant concepts from EventNet that can be used to search video corpus to answer a new event query. The existing methods address this by calculating the semantic similarity between the event query and the candidate concepts, and then picking up the top ranked concepts with the highest similarities~\cite{Chen:ICMR14, Cui:arXiv14,Wu:CVPR14}. However, considering that our concepts are event-specific, each concept is associated with a specific event which can be used as contextual information in measuring the similarity between the query event and the concept. Moreover, due to the short text of event names which contain only very few text words, direct measurement of semantic similarity may not be able to accurately estimate semantic relevance, and the concept matching results may become quite unsatisfactory even when EventNet library does contain relevant events and concepts. To solve these issues, we propose a cascaded concept matching method which first matches relevant events and then finds relevant concepts specific to the matched events. For the queries that cannot be well answered by automatic semantic similarity calculation, we propose to leverage the hierarchical structure of EventNet and allow users to manually specify the appropriate high-level category\footnote{As shown in Figure~\ref{Fig:Framework}, category nodes in EventNet are high-level categories in WikiHow used to organize articles into a hierarchy, such as ``pets and animals'', ``sports and fitness'', etc.} in the EventNet tree structure, and then only perform concept matching under the specified category (cf. Section~\ref{sec:semanticmatching})

We will demonstrate that the proposed EventNet concept library leads to dramatic performance gains in concept based event detection over various benchmark video event datasets. Specially, it outperforms the $20K$ concepts generated from the state-of-the-art deep learning system trained on ImageNet~\cite{Krizhevsky:NIPS12} by $207\%$ in zero-shot event retrieval. We will also show that EventNet is able to detect and recount the semantic cues indicating occurrence of an event video. Finally, the video corpus in EventNet can be used as a comprehensive benchmark video event dataset. The browser of EventNet ontology as well as the downloading information of the models and video data can be found at \url{http://eventnet.ee.columbia.edu}.

This paper presents several major contributions: (1) A systematic framework for discovering a large number of events related to human events (Section~\ref{Sec:ChooseEvent}). (2) Construction of the largest ontology including $500$ complex events and $4,490$ event-specific concepts (Section~\ref{Sec:EventNetConstruct} and~\ref{Sec:ModelTraining}). (3) Rigorous analysis of the properties of the constructed ontology (Section~\ref{Sec:EventAnalysis}). (4) Dramatic performance gains in complex event detection especially for unseen novel events (Task I in Section~\ref{Sec:Experiment}). (5) The benefit of the proposed ontology structure in semantic recounting (Task II in Section~\ref{Sec:Experiment}) and concept matching (Task III in Section~\ref{Sec:Experiment}). (6) A benchmark event video dataset for advancing large scale event detection (Task IV in Section~\ref{Sec:Experiment}).


\section{Related Work}
There are some recent works that focus on detecting video events using concept-based representations. 
For example, Wu~\emph{et al}.~\cite{Wu:CVPR14} mined concepts from the free-form text descriptions of TRECVID research video set, and applied them as weak concepts of the events in TRECVID MED task. As mentioned earlier, these concepts are not specifically designed for events, and may not capture well the semantics of event videos.

Recent research also attempts to define event-driven concepts for event detection. Liu~\emph{et al}.~\cite{Liu:WACV13} proposed to manually annotate related concepts in event videos, and build concept models with the annotated video frames. Chen \emph{et al}.~\cite{Chen:ICMR14} proposed to discover event-driven concepts from the tags of Flickr images crawled by using keywords of the events of interest. This method is able to find relevant concepts for each event and achieves good performance in various event detection tasks. Despite such promising properties, it heavily relies on the prior knowledge about the target events, and therefore cannot handle novel unknown events that may emerge at a later time. Our EventNet library tries to address this deficiency by exploring a large number of events and their related concepts from external knowledge resources, WikiHow and YouTube. A related prior work~\cite{Cui:arXiv14} tried to define a large number of events and discover concepts by using tags of Flickr images. However, as our later experiment will show, concept models trained with Flickr images cannot generalize well to event videos due to the well-known cross-domain data variation~\cite{Saenko:ECCV10}. In contrast, our method discovers concepts and trains models based on YouTube videos, which more accurately captures the semantic concepts underlying content of user generated videos.

The proposed EventNet also introduces a benchmark video dataset for large scale video event detection. Current event detection benchmarks typically only contain a small number of events. For example, in the well known TRECVID MED task~\cite{MED:NIST}, significant efforts have been made to develop an event video dataset containing $48$ events. Columbia Consumer Video (CCV) dataset~\cite{Jiang:ICMR11} contains $9,317$ videos of $20$ events. Such event categories may also suffer from data bias, and thus fail to provide general models applicable to unconstrained real-world events. In contrast, EventNet contains $500$ event categories and $95K$ videos, which covers different aspects of human daily life and is believed to be the largest event dataset up to date. Another recent effort also tries to build a large scale structured event video dataset containing $239$ events~\cite{Jiang:arXiv15}. However, it does not provide semantic concepts associated with specific events like those defined in EventNet.

\begin{table}[t]
\centering
\begin{tabular}{c||c|c|c|c|c}
  \hline
  Dataset & EM & PM & RE & NM & Total Class \#\\
  \hline\hline
  MED 10-14~\cite{MED:NIST} & 16 & 17 & 15 & 0 & 48 \\
  CCV~\cite{Jiang:ICMR11} & 6 & 5 & 8 & 1 & 20 \\
  Hollywood~\cite{Laptev:CVPR08} & 6 & 0 & 1 & 0 & 7 \\
  KTH~\cite{Laptev:ICCV03} & 4 & 1 & 1 & 0 & 6 \\
  UCF101~\cite{Soomro:CRCV12} & 58 & 11 & 20 & 12 & 101\\ \hline
  Matched Class \# & 90 &  34 &  45 & 13 & 182\\
  \hline
\end{tabular}\caption{The matching results between WikiHow articles and event classes in the popular event video datasets, where ``EM'' denotes ``Exact Match'', ``PM'' denotes ``Partial Match'', ``RE'' denotes ``Relevant'' and ``NM'' denotes ``No Match''. }\label{Tab:match_stats}
\end{table}


\section{Choosing WikiHow as EventNet Ontology}\label{Sec:ChooseEvent}
A key issue in constructing a large scale event driven concept library is to define an ontology that covers as many real-world events as possible. For this, we resort to the Internet knowledge bases constructed from crowd intelligence as our ontology definition resources. Specifically, WikiHow is an online forum that contains a large number of how-to manuals on every aspect of human daily life events, where a user can submit an article in describing how to accomplish a task such as ``how to bake sweet potatoes'', ``how to remove tree stumps'', etc. We choose WikiHow as our event ontology definition resource due to the following reasons: 

\textbf{Coverage of WikiHow Articles}. WikiHow has a good coverage over different aspects of human daily life events. As of February 2015, it included over $300K$ how-to articles~\cite{WikiHow}, among which some are well-defined video events\footnote{We define an event as video event when it satisfies the event definition in NIST TRECVID MED evaluation~\cite{MED:NIST}, i.e., a complicated human activity interacting with people/object in a certain scene.} that can be detected by computer vision techniques, while others such as ``how to think'' or ``how to apply for a passport'' do not have suitable corresponding video events. We expect a comprehensive coverage of video events from such a massive number of articles created by the crowdsourcing knowledge from Internet users.

\begin{figure}[t]
\centering
{\includegraphics[width=1\linewidth]{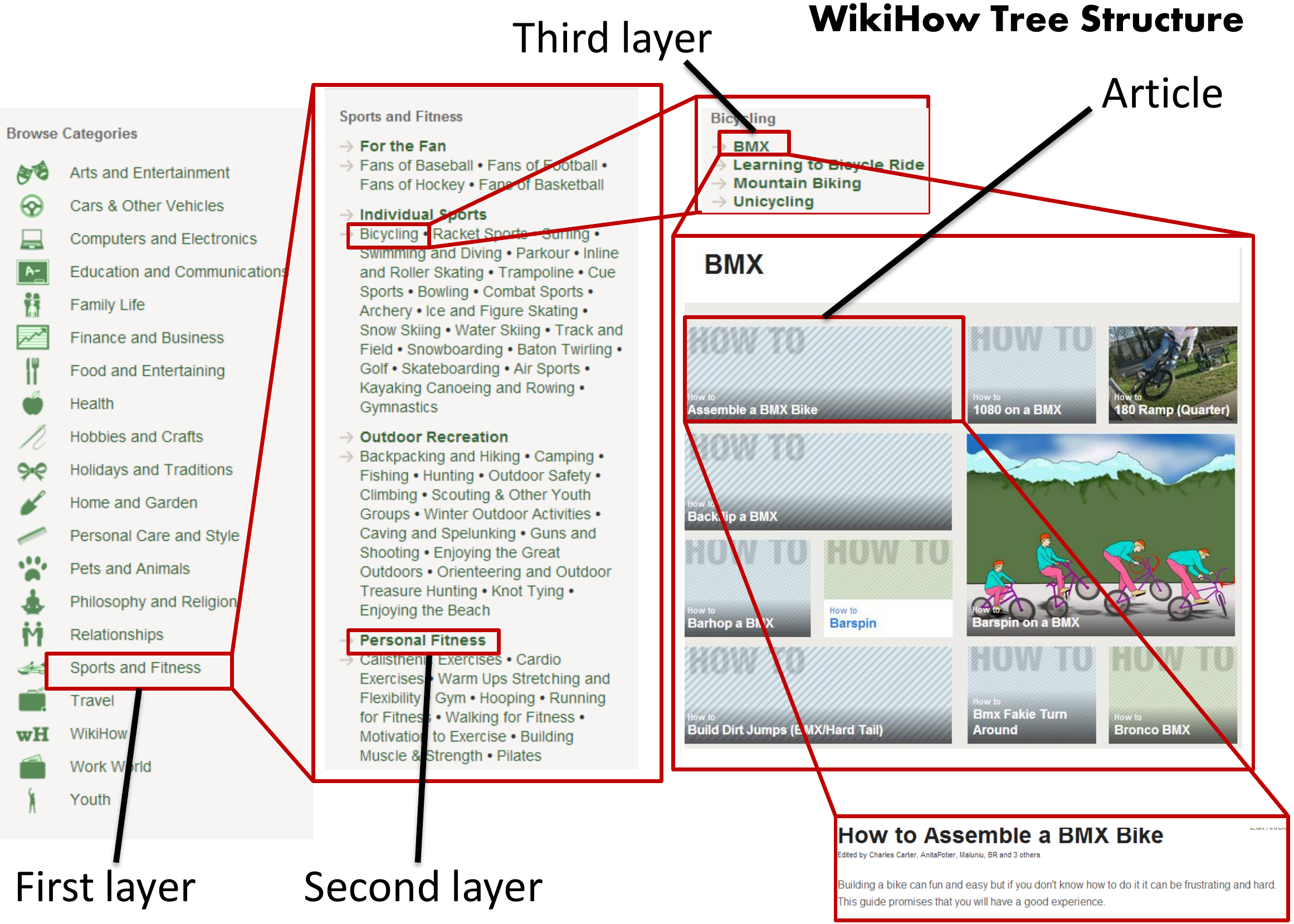}}
\caption{The hierarchial structure of WikiHow.}
\label{Fig:WikiHow}
\end{figure}

To verify that WikiHow articles have a good coverage on video events, we conduct a study to test whether WikiHow articles contain the events in the existing popular event video datasets in computer vision and multimedia fields.
To this end, we choose the event classes in the following datasets: TRECVID MED 2010-2014 ($48$ classes)~\cite{MED:NIST}, Columbia Consumer Videos (CCV) ($20$ classes)~\cite{Jiang:ICMR11}, UCF 101 ($101$ classes)~\cite{Soomro:CRCV12}, Hollywood movies ($7$ classes)~\cite{Laptev:CVPR08}, KTH ($6$ classes)~\cite{Laptev:ICCV03}. Then, we use each event class name as a text query to search WikiHow and examine the top-$10$ returned articles, from which we manually pick up the most relevant article title as the matching result. We define four matching levels to measure the matching quality. The first is \emph{exact matching}, in which the matched article title and the event query are exactly matched (e.g., ``clap hands'' as a matched result to query ``hand clapping''). The second is \emph{partial match}, in which the matched article is discussing about a certain case of the query (e.g., ``make a chocolate cake'' as result to query ``make a cake''). The third case is \emph{relevant}, in which the matched article is semantically relevant to the query (e.g., ``get your car out of the snow'' as result to query ``getting a vehicle unstuck''). The fourth case is \emph{no match}, in which we cannot find any relevant articles about the query. The matching statistics can be shown in Table~\ref{Tab:match_stats}. If we count the first three types of matching as successful cases, the coverage rate of WikiHow over these event classes is as high as $169/182=93\%$, which confirms the potential of discovering video events from WikiHow articles.  

\begin{figure}[htp]
\centering
{\includegraphics[width=1\linewidth]{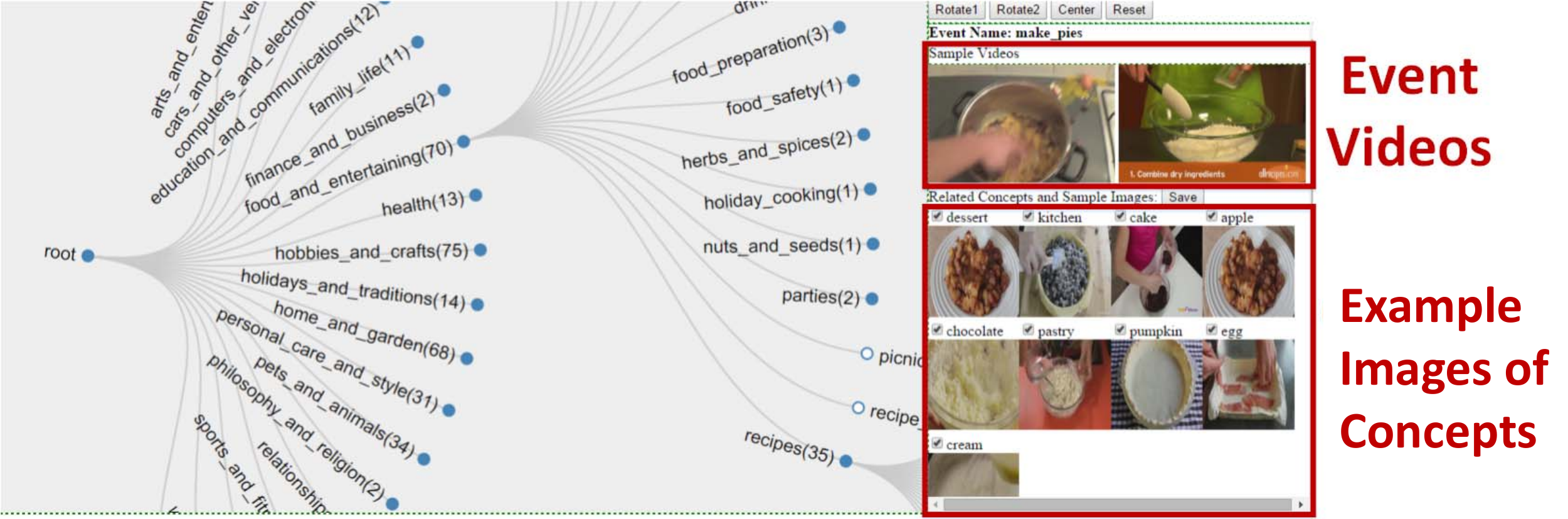}}\vspace{-2mm}
\caption{Event and concept browser for the proposed EventNet ontology. The hierarchical structure is shown on the left and the example videos and relevant concepts of each specific event are shown to the right.}
\label{Fig:Browser}
\end{figure}


\textbf{Hierarchical Structure of WikiHow}. WikiHow categorizes all its articles into $2,803$ categories and further organizes all categories into a hierarchical tree structure. Each category contains a number of articles discussing different aspects of the category, and is associated with a node in WikiHow hierarchy. As shown in Figure~\ref{Fig:WikiHow} of the WikiHow hierarchy, the first layer contains $19$ high-level nodes ranging from ``arts and entertainment'', ``sports and fitness'' to ``pets and animal''. Each node is further split into a number of children nodes that are subclasses or facets of the parent node with the deepest path from the root to the leaf node containing seven levels. Although such a hierarchy is not based on lexical knowledge, it summarizes human's common practice in organizing daily life events. Typically, a parent category node includes articles that are more generic than those in its children nodes. Therefore, the events residing along similar path in the WikiHow tree hierarchy are highly relevant (cf. Section~\ref{Sec:EventNetConstruct}). Such hierarchical structure will help users quickly localize the potential search area in the hierarchy for a specific query that he/she is interested in, and thus improve the concept matching accuracy (cf. Section~\ref{sec:semanticmatching}). Besides, such hierarchical structure will also enhance event detection performance by leveraging the detection result of an event in a parent node to boost detection of the events in its children nodes, and vice versa. Finally, such hierarchical structure also allows us to develop an intuitive browsing interface for event navigation and event detection result visualization~\cite{Xu:MM15}, which can be illustrated in Figure~\ref{Fig:Browser}.

\section{Constructing EventNet}\label{Sec:EventNetConstruct}
In this section, we describe the procedure used in constructing EventNet, including how to define video events from WikiHow articles and how to discover event specific concepts for each of the events from tags of YouTube videos.

\subsection{Discovering Events}
First, we aim to discover potential video events from WikiHow articles. Intuitively, this can be done by crawling videos using each article title and then applying the automatic verification technique as that proposed in~\cite{Berg:ECCV10,Chen:ICMR14} to determine whether an article corresponds to a video event. However, considering that there are $300K$ articles on WikiHow, it requires massive amount of data crawling and video processing, making it computationally infeasible. For this, we propose a coarse-to-fine event selection approach. The basic idea is to first prune WikiHow categories that do not correspond to video events, and then select one representative event from the article titles within each of the remaining categories. In the following, we will describe the event selection procedure in details. 

\textbf{Step I: WikiHow Category Pruning}. Recall that WikiHow contains $2,803$ categories, each of which contains a number of articles about this category. We observe that many of the categories are talking about personal experience and suggestions, which do not correspond to video events. For example, the articles in the category of ``Living Overseas'' are all talking about how to improve living experience in a foreign country and do not satisfy the definition of video event. Therefore, we want to find out such event irrelevant categories and then directly filter out their articles, so that we can significantly prune the number of articles to be verified in the next stage. To this end, we go through $2,803$ WikiHow categories and manually remove those that are irrelevant to video events. A category is deemed as event irrelevant when it cannot be visually described by a video, and none of its articles contain any video events. For example, ``Living Overseas'' is an event-irrelevant category since ``Living Overseas'' is not visually observable in videos and none of its articles are events. On the other hand, although category ``Science'' cannot be visually described in a video due to its abstract meaning, it contains some instructional articles corresponding to video events, such as ``Make Hot Ice'', and ``Use a Microscope''.  As a result, in our manual pruning procedure, we first find a to-be-pruned category name and then carefully review their articles before we decide to remove this category.

\textbf{Step II: Category Based Event Selection}. After category pruning, only event relevant categories and their articles remain. Under each category, there are still a large number of articles that are not corresponding to events. Our eventual goal is to find all video events from these articles and include them into our event collection, which is a long term goal of EventNet project. In the current version, EventNet only includes one representative video event from each category of WikiHow ontology. An article title is considered to be a video event when it satisfies the following four conditions: (1) It defines an event that involves human activity interacting with people/objects in a certain scene. (2) It has concrete non-subjective meanings. For example, ``decorating a romantic bedroom'' is too subjective since different users may have different interpretation about ``romantic''. (3) It has consistent observable visual characteristics. For example, a simple method is to use the candidate event name to search YouTube and check whether there are consistent visual tags found in the top returned videos. Tags may be approximately considered visual if they can be found in an existing image ontology such as ImageNet. (4) It is generic, not too detailed. If many article titles under a category share the same verb and direct object, then they can be formed to a generic event name. After this, we end up with $500$ event categories as the current event collection in EventNet. 

\begin{figure}[t]
\centering
{\includegraphics[width=1\linewidth]{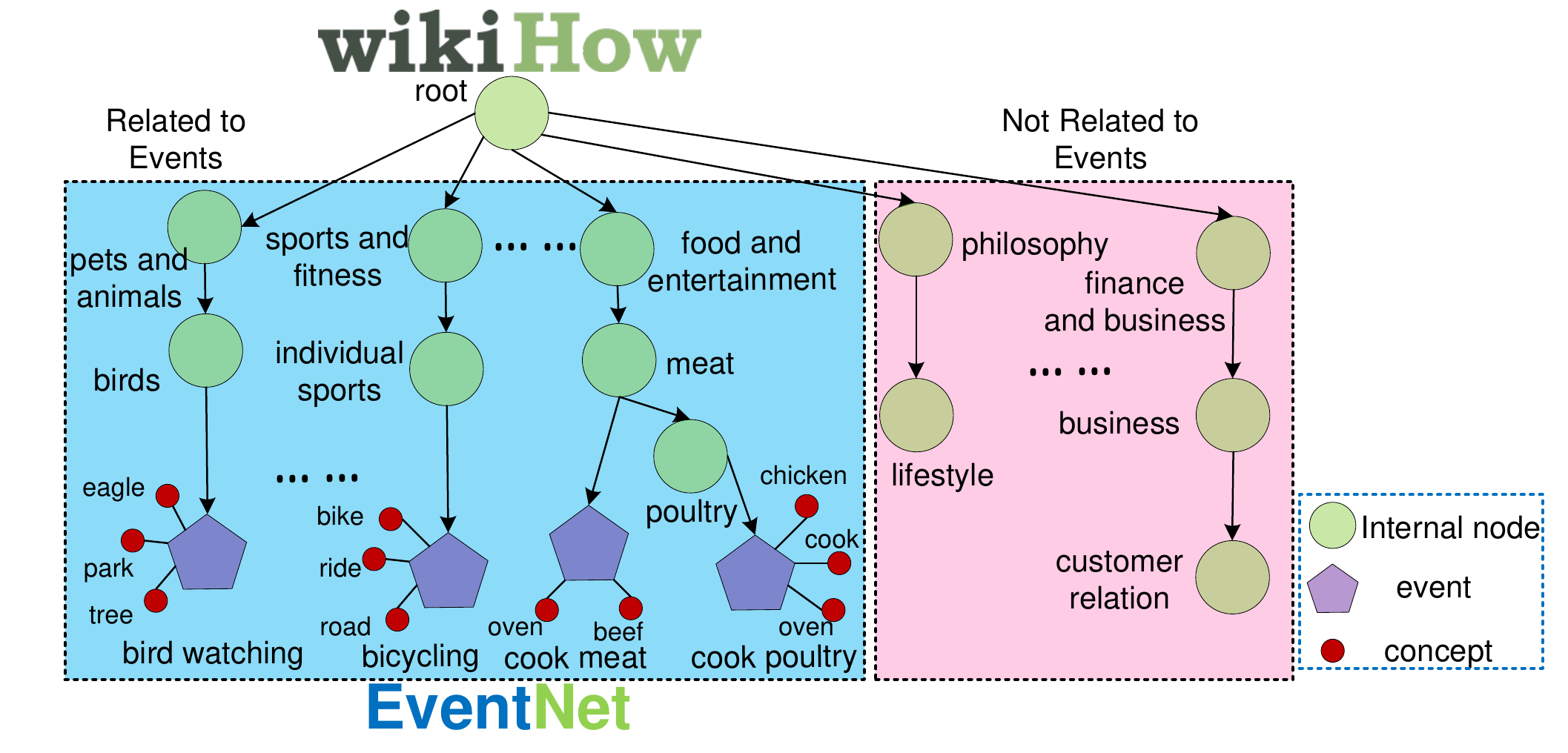}}
\caption{A snapshot of EventNet constructed from WikiHow.}
\label{Fig:EventNet}
\end{figure}

\subsection{Mining Event Specific Concepts}
We apply the concept discovery method developed in our prior work~\cite{Chen:ICMR14} to discover event-driven concepts from the tags of YouTube videos. For each of the $500$ events, we use the event
name as query keywords to search YouTube. We check the top $1,000$ returned videos and collect the $10$ most frequent words appearing in the titles or tags of these videos. Then we further filter the $1,000$ videos to only include videos containing at least three of the frequent words. This step helps us remove many irrelevant videos from the searching results. Using this approach, we crawl around $190$ videos and their tag lists as concept discovery resource for each event, ending up with $95,321$ videos for $500$ events. We discover event-specific concepts from the tags of the crawled videos. To ensure the visual detectability of the discovered concepts, we match each tag to the classes of the existing object (ImageNet~\cite{Deng:CVPR10}), scene (SUN~\cite{Patterson:CVPR12}) and action (Action Bank~\cite{Sadanand:CVPR12}) libraries, and only keep the matched words as the candidate concepts. After going through the process, we end up with around nine concepts per event, and in total $4,490$ concepts for the entire set of $500$ events. Finally, we adopt the hierarchical structure of WikiHow categories and attach each discovered event and its concepts to the corresponding category node. The final event concept ontology is called EventNet, as illustrated in Figure~\ref{Fig:EventNet}.

One may argue that the construction of EventNet ontology heavily depends on subjective evaluation. In fact, we can replace such subjective evaluation with automatic methods from computer vision and natural language processing techniques. For example, we can use concept visual verification to measure the visual detectability of concepts~\cite{Chen:ICMR14}, and use text based event extraction to determine whether each article title is an event~\cite{Ritter:KDD12}. However, as the accuracy of such automatic methods is still being improved, currently we focus on design of principled criteria for event discovery and defer incorporation of automatic discovery processes until future improvement.

\begin{figure}[hpt]
\centering
{\includegraphics[width=1\linewidth]{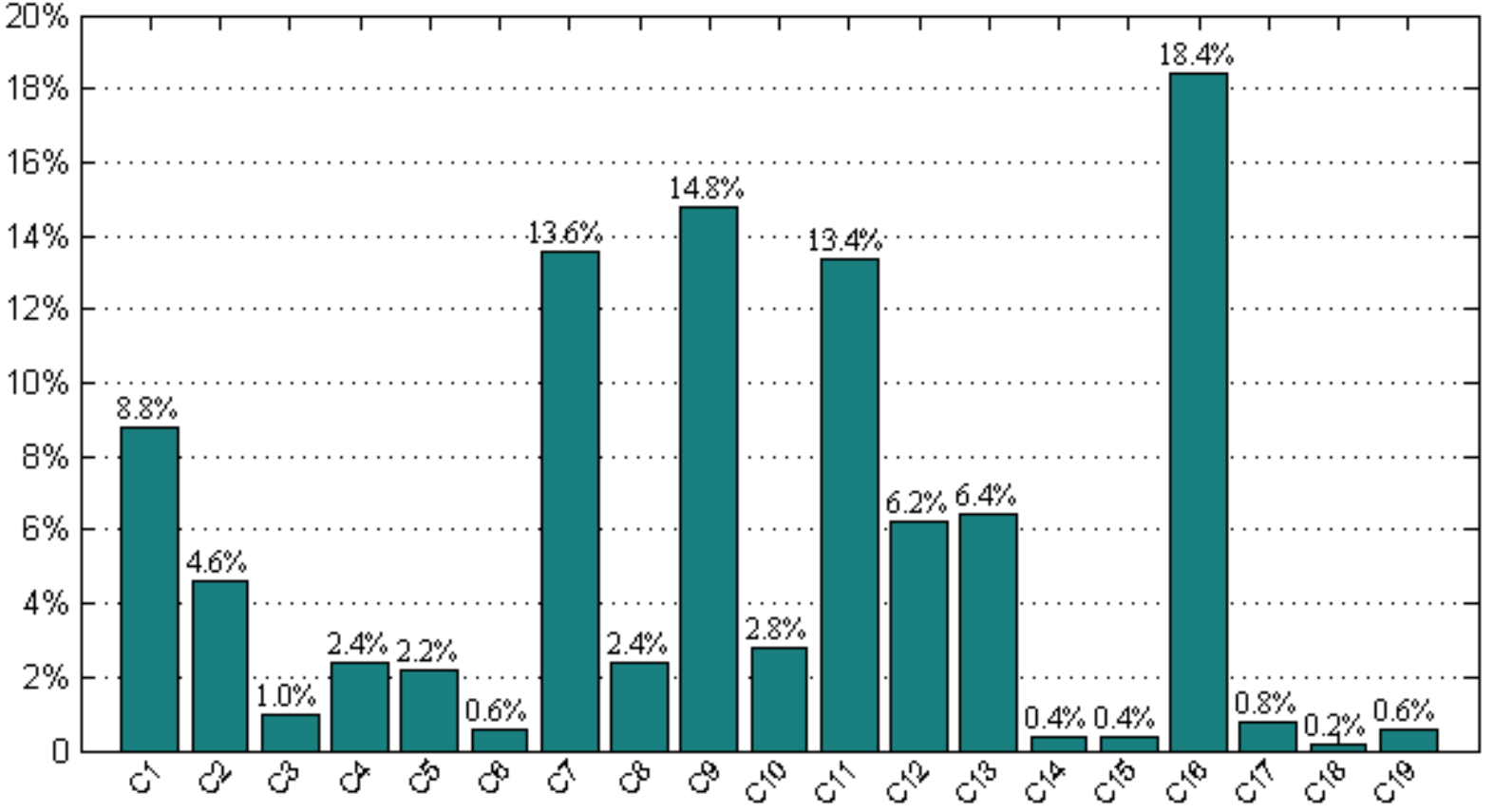}}
\caption{Event distribution over the top-$19$ categories of EventNet, where C1 to C19 are ``arts and entertainment'', ``cars and other vehicles'', ``computers and electronics'', ``education and communications'', ``family life'', ``finance and business'', ``food and entertaining'', ``health'', ``hobbies and crafts'', ``holidays and traditions'', ``home and garden'', ``personal care and style'', ``pets and animals'', ``philosophy and religion'', ``relationships'', ``sports and fitness'', ``travel'', ``work world'', and ``youth''.}
\label{Fig:Distribution}
\end{figure}

\section{Properties of EventNet}\label{Sec:EventAnalysis}
In this section, we will provide detailed analysis on the properties of EventNet ontology, including basic statistics about the ontology, event distribution over coarse categories, and event redundancy. 

\textbf{EventNet Statistics}. EventNet ontology contains $682$ WikiHow category nodes, $500$ event nodes and $4,490$ concept nodes organized in a tree structure, in which the deepest depth from the root node to the leaf node (the event node) is eight. Each non-leaf category node has four child category nodes in average. Regarding the video statistics in EventNet, the average number of videos per event is $190$, and the number of videos per concept is $21$. EventNet has $95,321$ videos with an average duration of around $277$ seconds ($7,334$ hours in total). 

\textbf{Event Distribution}. We show the percentage of the number of events distributed over the top-$19$  category nodes of EventNet, and the results can be shown in Figure~\ref{Fig:Distribution}. As seen, the top four popular categories including the most number of events are ``sports and fitness'', ``hobbies and craft'', ``food and entertainment'', and ``home and garden'', while the least four populated categories are `` work world'', ``relationships'', ``philosophy and religion'' and ``youth'', which are abstract and cannot be described in videos.  A further glimpse of the event distributions tells us that the most popular categories reflects user's common interests in video content creation. For example, most event videos captured in human daily life are talking about their life styles reflected in food, fitness, and hobbies.  Therefore, we believe that events included in EventNet have the potential to be used as an event concept library to detect popular events in human daily life. 

\textbf{Event Redundancy}. We also carry out analysis on the redundancy among the $500$ events in EventNet. To this end, we use each event name as a text query, and find its most semantically similar events from other events located at different branches from the query event. Specifically, given a query event $e_q$, we first localize its category node $C_q$ in EventNet tree structure, and then exclude all events attached under the parent nodes and children nodes of node $C_q$. The events attached to other nodes are treated as the search base to find similar events of the query based on the semantic similarity described in Section~\ref{sec:semanticmatching}. The reason on excluding events at the same branch of the query event is due to the fact that events residing at parent category nodes and children category nodes manifest hierarchical relations such as ``cook meat'' and ``cook poultry''. We treat such hierarchical event pairs as a desired property of EventNet library, and therefore do not involve them into redundancy analysis. From the top-$5$ ranked events for a given query, we ask human annotators to determine whether there is a redundant event talking about the same event as the query. After applying all $500$ events as queries, we find zero redundancy among query event and all other events residing at different branches of EventNet structure. 

\begin{table*}[htp]
\centering
\begin{tabular}{c||c|c}
\hline
Event Query & without leveraging EventNet structure & with leveraging EventNet structure \\
\hline \hline
\multirow{2}{*}{\textbf{\tabincell{c}{landing a fish}}} & landing a plane & fishing \\
 & cook fish & hunt an animal  \\
\hline
\multirow{2}{*}{\textbf{\tabincell{c}{wedding shower}}} & wedding ceremony & wedding ceremony \\
 & take a shower & make a wedding veil  \\
\hline
\multirow{2}{*}{\textbf{\tabincell{c}{working on a woodworking project}}} & working out using a rowing machine &  make wood projects\\
 & running & make a crochet project  \\
\hline

\end{tabular}
\caption{Top $2$ matched events of some event queries without ($2$nd column) and with ($3$rd column) leveraging EventNet structure.}
\label{Tab:EventMatch}
\end{table*}

\section{Learning Concept Models from Deep Learning Video Features}\label{Sec:ModelTraining}
In this section, we introduce the procedure of learning concept classifiers for EventNet concept library. Our learning framework leverages the recent powerful CNN model to extract deep learning feature from the video content, while employing one-vs-all linear SVM trained on top of the features as concept models.

\subsection{Deep Feature Learning with CNN}\label{Sec:CNN}
We adopt the CNN architecture in~\cite{Krizhevsky:NIPS12} as deep learning model to perform deep feature learning from the video content. The network takes RGB video frame as input and outputs the score distribution over the $500$ events in EventNet.
The network has five successive convolutional layers followed by two fully connected layers. The detailed information about the network architecture can be found in~\cite{Krizhevsky:NIPS12}. In this work, we apply \emph{Caffe}~\cite{Jia:Caffe13} as the implementation of the CNN model described by~\cite{Krizhevsky:NIPS12}.

For the training of EventNet CNN model, we evenly sample $40$ frames from each video, and end up with four millions frames over all $500$ events as the training set. For each of the $500$ events, we treat the frames sampled from its videos as the positive training samples of this event. We define the set of $500$ events as $E=\{0,1,...,499\}$. Then the prediction probability of $k$-th event for input sample $n$ is defined as:

\begin{equation}
    p_{nk}=\frac{\exp(x_{nk})}{\sum_{k'\in E}\exp(x_{nk'})},
    \label{eq1}
\end{equation}
where $x_{nk}$ is the $k$-th node's output of the $n$-th input sample from CNN's last layer. The loss funtion $L$ is defined as a multinomial logistic loss of the softmax which is $L=\frac{-1}{N}\sum^N_{n=1}\log(p_{n,l_n})$, where $l_n\in E$ indicating the correct class label for input sample $n$ and $N$ is the total number of inputs. Our CNN model is trained on NVIDIA Tesla K20 GPU, and it takes about seven days to finish $450K$ iterations of training.  After CNN training, we extract the $4,096$ dimensional feature vector from the second to the last layer of the CNN architecture, and further perform $\ell_2$ normalization on the feature vector as the deep learning feature descriptor of each video frame.

\subsection{Concept Model Training}
Given a concept discovered for an event, we treat the videos associated with this concept as positive training data and randomly sample the same number of videos from concepts in other events as negative training data. This obviously has the risk of generating false negatives (videos without a certain concept label does not necessarily mean it is negative for the concept). But in view of the prohibitive cost in annotating all videos over all concepts, we follow this common practice used in other image ontologies such as ImageNet~\cite{Deng:CVPR10}. We directly treat frames in positive videos as positive and frames in negative videos as negative to train a linear SVM classifier as the concept model.
%
This is a simplified approach and there are emerging works~\cite{Lai:ECCV14} in selecting more precise temporal segments or frames in videos as positive samples.

To generate concept scores on a given video, we first uniformly sample frames from it and extract the $4,096$ dimensional CNN feature from each frame. Then we apply the $4,490$ concept models on each frame and use all $4,490$ concept scores as the concept representation of this frame. Finally, we average the score vectors across all frames and adopt the average score vector as the video level concept representation.

\section{Leveraging EventNet Structure for Concept Matching}\label{sec:semanticmatching}
In concept based event detection, the first step is to find some semantically relevant concepts that are applicable for detecting videos with respect to the event query.
This procedure is called \emph{concept matching} in the literature~\cite{Chen:ICMR14,Wu:CVPR14}. To accomplish this task, the existing approaches typically calculate the semantic similarity between the query event and each concept in the library based on the external semantic knowledge bases such as WordNet~\cite{Fellbaum:BB98} or ConceptNet~\cite{Liu:BT04}, and then pick up the top ranked concepts as the relevant concepts for event detection. However, these approaches may not be applicable for our EventNet concept library since the involved concepts are event specific concepts dependent on their associative events. For example, concept ``dog'' under ``feed a dog'' and ``groom a dog'' will be treated as two different concepts due to different event contexts. Therefore, concept matching in EventNet needs to take into account event contextual information.

To this end, we propose a multi-step concept matching approach that first finds relevant events and then chooses concepts from the concepts associated with the matched events. Specifically, given an event query $e_q$ and an event $e$ in EventNet library, we use the textual phrase similarity calculation function developed in~\cite{Han:ACL13} to estimate their semantic similarity. The reason to adopt such semantic similarity function is due to the fact that both event query and candidate events in EventNet library are textual phrases, which need sophisticated phrase level similarity calculation supporting word sequence alignment and strong generalization ability achieved by machine learning. However, these properties cannot be achieved by using the standard similarity computation methods based on WordNet or ConceptNet alone. Our empirical studies confirm that the phrase based semantic similarity is able to obtain better event matching results.

However, due to the word sense ambiguity and the limited amount of text information in event names, the phrase similarity based matching approach can also generate wrong matching results. For example, given a query ``wedding shower '', event ``take a shower'' in EventNet will receive a high similarity value since ``shower'' has an ambiguous meaning, and thus be mistakenly matched as a relevant event. Likewise, the best matching results for query ``landing a fish'' are ``landing an airplane'' and ``cook fish'' rather than ``fishing'' which is the most relevant. To address these problems, we propose to exploit the structure of EventNet ontology to find relevant events for such difficult query events. Specifically, given the query event, users can manually specify the suitable categories in the top level of EventNet structure. For instance, users can easily specify that the suitable categories for event ``wedding shower'' is ``Family Life'' while choosing ``Sports and Fitness'' and ``Hobbies and Crafts'' as suitable categories for ``landing a fish''.  After user's specification, the subsequent event matching only needs to be conducted over the events under the specified high-level categories. In this way, the hierarchical structure of EventNet ontology will be helpful in relieving the limitations of short text based semantic matching, and help improve concept matching accuracy. Table~\ref{Tab:EventMatch} shows some difficult events from TRECVID MED and their top matched events with and without leveraging EventNet structure. As seen, our method can achieve more relevant matching results than using phrase based semantic similarity alone. After we get the top matched events, we can further choose concepts based on their semantic similarity to the query event. Quantitative evaluations between the matching methods can be found in Section~\ref{Sec:StructureEval}.

\begin{figure*}[htp]
\centering
\subfigure
{\label{fig_sr:a}
\includegraphics[width=1\columnwidth]{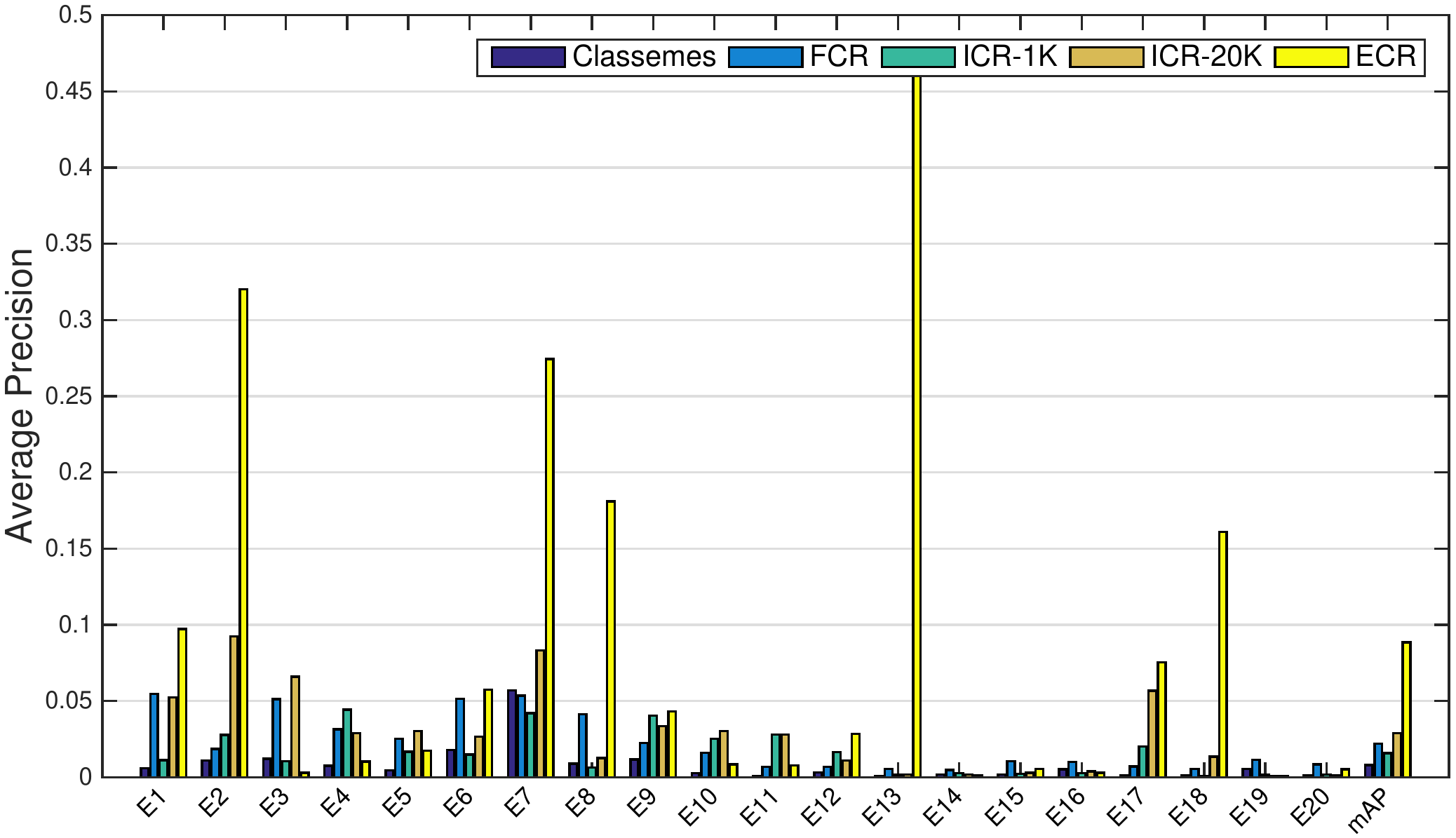}
}
\subfigure
{ \label{fig_sr:b}
\includegraphics[width=0.98\columnwidth]{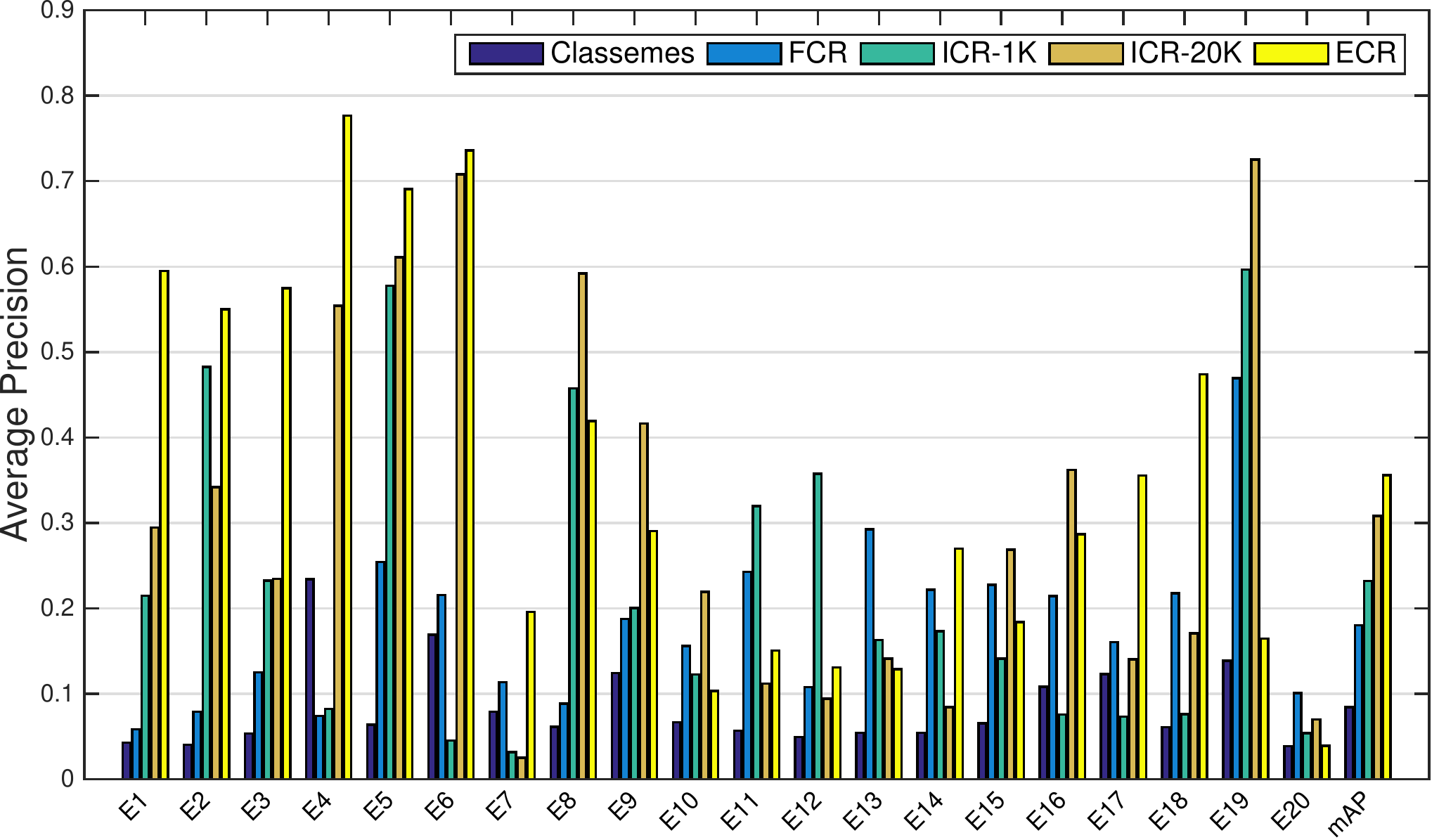}
}\vspace{-2mm}
\caption{Performance comparisons on zero-shot event retrieval task (left: MED; right: CCV). This figure is best viewed in color.}
\label{fig_sr}
\end{figure*}


\begin{figure*}[hpt]
\centering
\subfigure
{\label{fig_sr_noc:a}
\includegraphics[width=0.76\columnwidth]{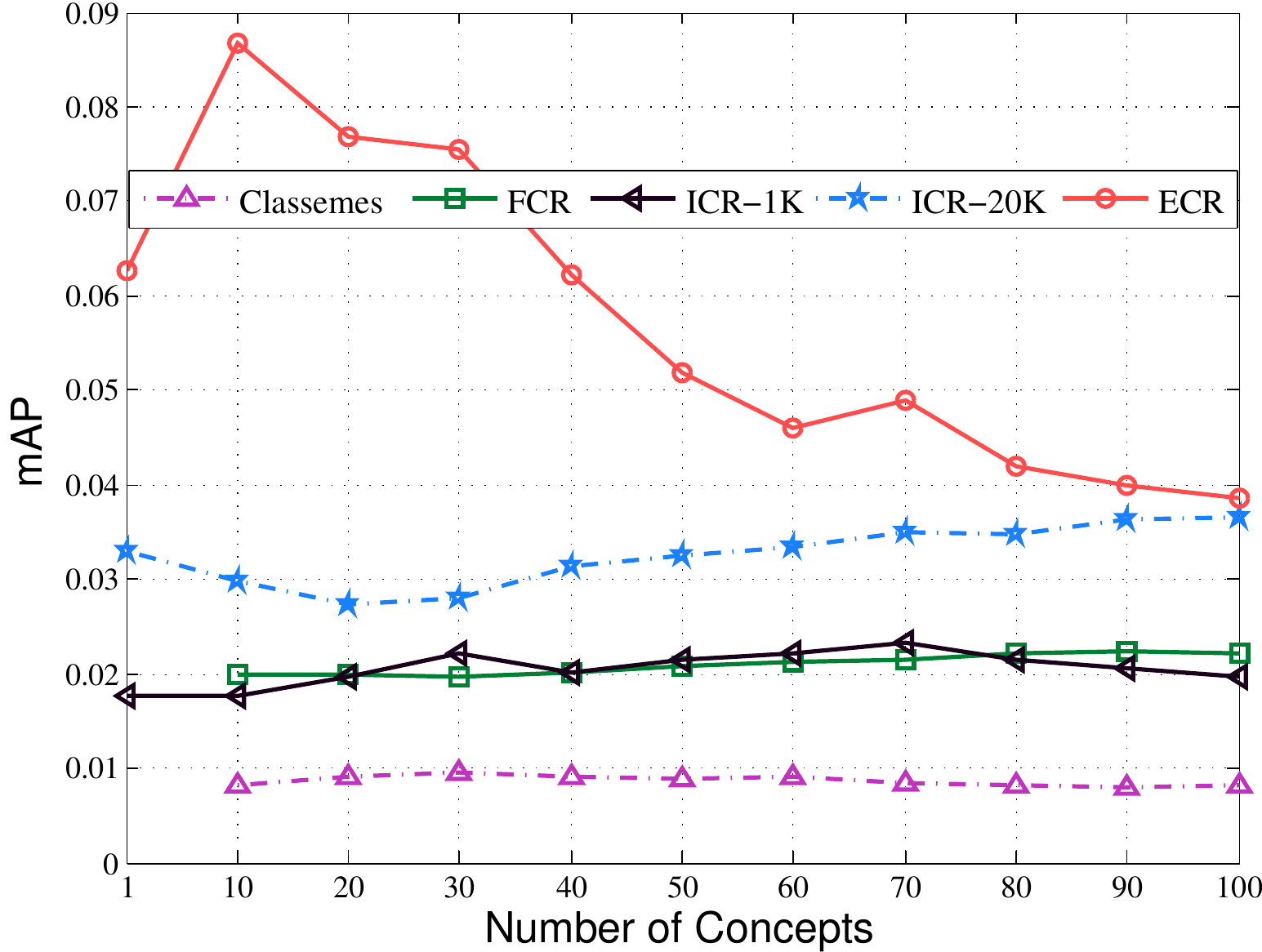}
}
\subfigure
{ \label{fig_sr_noc:b}
\includegraphics[width=0.76\columnwidth]{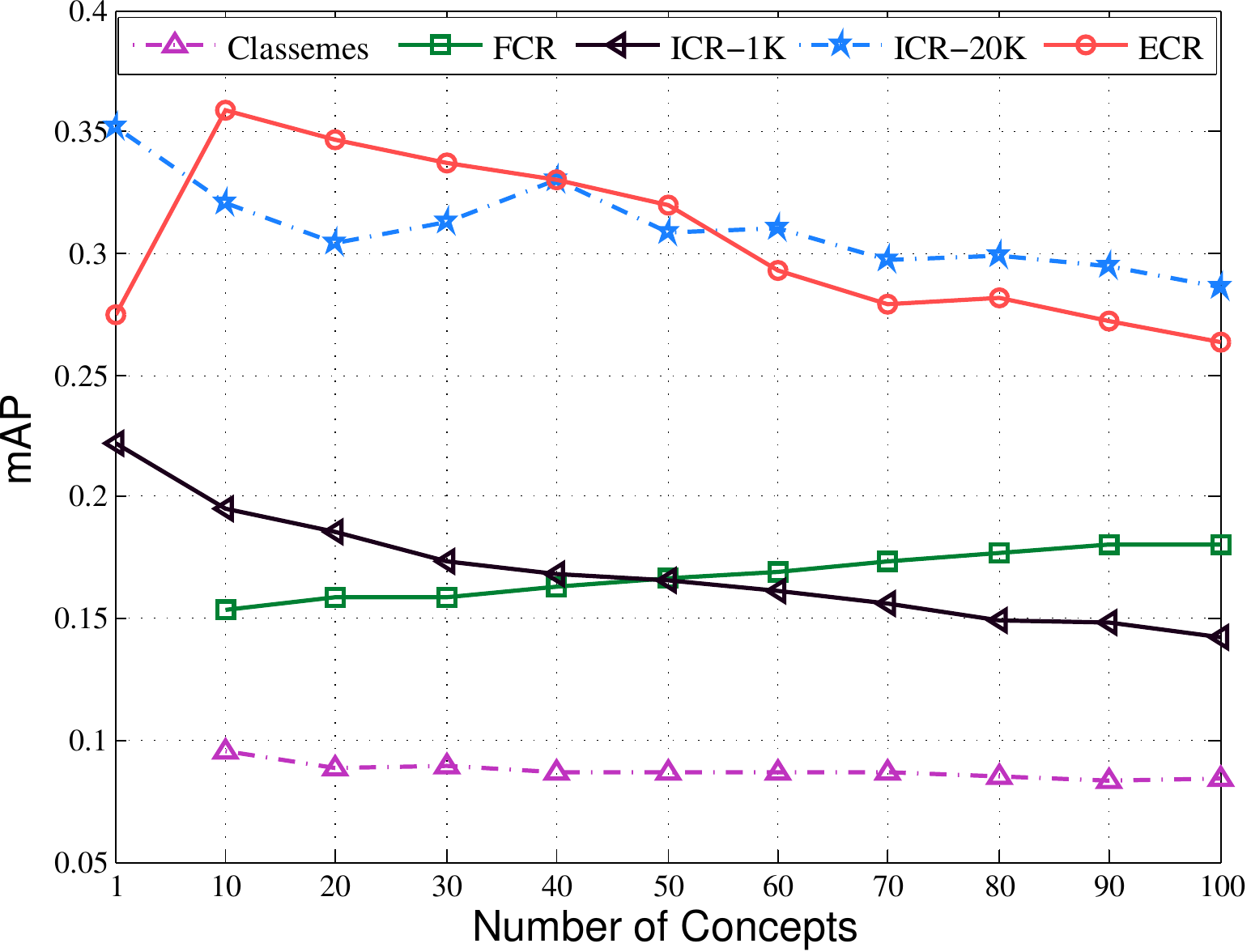}
}\vspace{-2mm}
\caption{Zero-shot event retrieval performance with different number of concepts (left: MED; right: CCV). The results of Classemes and FCR are from~\cite{Cui:arXiv14}, in which the results when concept number is $1$ are not reported.}
\label{fig_sr_noc}
\end{figure*}

\section{Experiments}\label{Sec:Experiment}
In this section, we evaluate the effectiveness of EventNet concept library in concept based event detection. We will first introduce the dataset and experimental setup, and then report the performance of different methods in the context of various event detection tasks, including zero-shot event retrieval and semantic recounting. After this, we will study the efforts of leveraging EventNet structure for matching concepts in zero-shot event retrieval. Finally, we will treat the $95K$ videos over $500$ events in EventNet as a video event benchmark and report the baseline performance of using the CNN model in event detection.

\subsection{Dataset and Experiment Setup}
\noindent\textbf{Dataset}. We use two benchmark video event datasets as the test sets of our experiments to verify the effectiveness of EventNet concept library. (1) \textit{TRECVID 2013 Multimedia Event Detection (MED)} dataset~\cite{MED:NIST}. It contains $32,744$ videos spanning over $20$ event classes and the distracting background, whose names are ``E1: \emph{birthday party}'', ``E2: \emph{changing a vehicle tire}'', ``E3: \emph{flash mob gathering}'', ``E4: \emph{getting a vehicle unstuck}'', ``E5: \emph{grooming an animal}'', ``E6: \emph{making a sandwich}'', ``E7: \emph{parade}'', ``E8: \emph{parkour}'', ``E9: \emph{repairing an appliance}'', ``E10: \emph{working on a sewing project}'', ``E11: \emph{attempting a bike trick}'', ``E12: \emph{cleaning an appliance}'', ``E13: \emph{dog show}'', ``E14: \emph{giving directions to a location}'', ``E15: \emph{marriage proposal}'', ``E16: \emph{renovating a home}'', ``E17: \emph{rock climbing}'', ``E18: \emph{town hall meeting}'', ``E19: \emph{winning a race without a vehicle}'', ``E20: \emph{working on a metal crafts project}''. We follow the original partition of this dataset in TRECVID MED evaluation, which partitions the dataset into a training set with $7,787$ videos and a test set with $24,957$ videos. (2) \textit{Columbia Consumer Video (CCV)} dataset~\cite{Jiang:ICMR11}. It contains $9,317$ videos spanning over $20$ classes, which are ``E1: \emph{basketball}'', ``E2: \emph{baseball}'', ``E3: \emph{soccer}'', ``E4: \emph{ice skating}'', ``E5: \emph{skiing}'', ``E6: \emph{swimming}'', ``E7: \emph{biking}'', ``E8: \emph{cat}'', ``E9: \emph{dog}'', ``E10: \emph{bird}'', ``E11: \emph{graduation}'', ``E12: \emph{birthday}'', ``E13: \emph{wedding reception}'', ``E14: \emph{wedding ceremony}'', ``E15: \emph{wedding dance}'', ``E16: \emph{music performance}'', ``E17: \emph{non-music performance}'', ``E18: \emph{parade}'', ``E19: \emph{beach}'', ``E20: \emph{playground}''. The dataset is further split into $4,659$ training videos and $4,658$ test videos. Since we focus on zero-shot event detection, we do not use the training videos in these two datasets, but only test the performance on the test set. For supervised visual recognition, features from deep learning models, e.g., the last few layers of deep learning models learned over ImageNet $1K$ or $20K$) can be directly used to detect events~\cite{Jain:CVPRW14}. However, the focus of this paper is on the semantic description power of the event-specific concepts, especially in recounting the semantic concepts in event detection and finding relevant concepts for retrieving events that have not been seen before (zero-shot retrieval).

\noindent\textbf{Feature Extraction}. On the two evaluation event video datasets, we extract the same feature as we did on EventNet videos. Specifically, we sample one frame every two seconds from a video, and extract the $4,096$ dimensional deep learning feature from the CNN model trained on EventNet video frames. Then we run SVM-based concept models over each frame, and aggregate the score vectors in a video as the semantic concept feature of the video.

\noindent\textbf{Comparison Methods and Evaluation Metric}. We compare different concept based video representations produced by the following methods. (1) \textbf{Classemes}~\cite{Torresani:ECCV10}. It is a $2,659$-dimensional concept representation whose concepts are defined based on LSCOM concept ontology. We directly extract Classemes on each frame and then average them across the video as video-level concept representation. (2) Flickr Concept Representation (\textbf{FCR})~\cite{Cui:arXiv14}. For each event, the concepts are automatically discovered from the tags of Flickr images in the search results of event query and organized based on WikiHow ontology. The concept detection models are based on the binary multiple kernel linear SVM classifiers trained with the Flickr images associated with each concept. Five kinds of low-level features are adopted to represent Flickr images and event video frames. (3) ImageNet-$1K$ CNN Concept Representation (\textbf{ICR-1K}). In this method, we directly apply the network architecture in~\cite{Krizhevsky:NIPS12} to train a CNN model over $1.2$ million high-resolution images in the ImageNet LSVRC-2010 contest that covers $1,000$ different classes~\cite{Russakovsky:arXiv14}. After model training, we apply the CNN model on the frames from both TRECVID MED and CCV datasets. Concept scores of the individual frames in a video are averaged to form the concept scores of the video. We treat the $1,000$ output scores as the concept based video representation from ImageNet-$1K$.  (4) ImageNet-$20K$ CNN Concept Representation (\textbf{ICR-20K}). We apply the same network architecture as ICR-1K to train a CNN model over $20$ million images spanning over $20,574$ classes from the latest release of ImageNet~\cite{Deng:CVPR10}. We treat the $20,574$ concept scores output from CNN model as the concept representation. Notably, ICR-1K and ICR-20K represent the most successful visual recognition achievements in computer vision area, which can be applied to justify the superiority of our EventNet concept library over the state-of-the-art. (5) Our proposed EventNet-CNN Concept Representation (\textbf{ECR}), in which we use our EventNet concept library to generate concept based video representations. (6) Some state-of-the-art results reported in the literature. Regarding the evaluation metric, we adopt Average Precision (AP), which approximates the area under precision/recall curve, to measure the performance on each event in our evaluation datasets. Finally, we calculate mean Average Precision (mAP) over all event classes as the overall evaluation metric.

\subsection{Task I: Zero-Shot Event Retrieval}
Our first experiment will evaluate the performance of zero-shot event retrieval, in which we do not use any training videos, but completely depend on the concept scores on test videos. To this end, we use each event name in the two video datasets as query to match two most relevant events and choose $15$ most relevant EventNet concepts based on semantic similarity, and then average the scores of these $15$ concepts as the zero-shot event detection score of video, through which a video ranking list can be generated.  Notably, the two most relevant events mentioned  above are automatically selected based on the semantic similarity matching method described in Section~\ref{sec:semanticmatching}.  For Classemes and FCR, we follow the setting in~\cite{Cui:arXiv14} to choose $100$ relevant concepts based on semantic similarity using ConceptNet and the concept matching method described in~\cite{Cui:arXiv14}. For ICR-1K and ICR-20K, we choose $15$ concepts using the same concept matching method.

Figure~\ref{fig_sr} shows the performance of different methods on two datasets respectively. From the results, we have the following observations:
(1) Event specific concept representations including FCR, ECR outperform the event independent concept representation Classemes. This is due to the fact that the former not only discovers semantically relevant concepts of the event, but also leverages the contextual information about the event in the training samples of each concept. In contrast, the latter only borrows concepts that are not specifically designed for events, and the training images for concept classifiers do not contain the event-related contextual information. (2) Concept representations trained with deep CNN features, including ICR-20K and ECR, produce much higher performance than the concept representations learned from low-level features including Classemes and FCR for most of the events. This is reasonable since CNN model can extract learning based features that have been shown to achieve strong performance. (3) Although all are trained with deep learning features, ECR generated by our proposed EventNet concept library performs significantly better than ICR-1K and ICR-20K, which are generated by deep learning models trained on ImageNet images. The reason is that concepts in EventNet are more relevant to events than the concepts in ImageNet which are mostly objects independent of events. From this result, we can see that our EventNet concepts even beat the concepts from the state-of-the-art visual recognition system, and is believed to be a powerful concept library for the task of zero-shot event retrieval.

Notably, our ECR achieves significant performance gains over the best baseline ICR-20K where the mAP on TRECVID MED increases from $2.89\%$ to $8.86\%$ with $207\%$ relative improvement. Similarly, the mAP on CCV increases from $30.82\%$ to $35.58\%$ ($15.4\%$ relative improvement). Moreover, our ECR achieves the best performances on most of the event categories on each dataset. For instance, on event ``E02: \emph{changing a vehicle tire}'' of TRECVID MED dataset, our method outperforms the best baseline ICR-20K by $246\%$ relative improvement. On TRECVID MED dataset, the reason of large improvement on ``E13: \emph{dog show}'' is because the matched events contain exactly the same event ``dog show'' as the event query. The performance on E10 and E11 is not so good since the automatic event matching method matched them to wrong events. When we use EventNet structure to correct the matching errors as described in Section~\ref{Sec:StructureEval}, we achieve higher performance on these events.

In Figure~\ref{fig_sr_noc}, we show the impact on zero-shot event retrieval performance when the number of concepts changes by using the concept matching method described in Section~\ref{sec:semanticmatching}, i.e., we first find the matched events, and then pick up the top ranked concepts belonging to these events. We select the number of events until the desired number of concepts is reached. On TRECVID MED, we can see consistent and significant performance gains of our proposed ECR method over others. However, on CCV dataset, ICR-20K achieves similar or even better performance when a large number of concepts is adopted. We conjecture this is due to the fact that CCV dataset contains a number of object categories such as ``E8: cat'' and ``E9: dog'', which might be better described by the visual objects contained in ImageNet dataset. Alternatively, all events in TRECVID MED are highly complicated events, which may be better described by EventNet. It is worth mentioning that the mAP first increases and then decreases as we choose more concepts from EventNet. This is because our concept matching method always ranks the most relevant concepts on top of the concept list. Therefore, involving many less relevant concepts ranked at lower positions (after the $10$th position in this experiment) in the concept list may decrease the performance. In Table~\ref{Tab:ComWithStateArt}, we compare our result with the state-of-the-art results reported on TRECVID MED 2013 test set with the same experiment setting. We can see that our ECR method outperforms these results by a large margin.

\begin{table}[htp]
\begin{center}
\begin{tabular}{|l|c|}
\hline
Method & mAP ($\%$)\\
\hline\hline
Selective concept~\cite{Habibian:ICMR14,Mazloon:ICMR13} & $4.39$\\
Bi-concept~\cite{Habibian:ICMR14,Rastegari:CVPR13} & $3.45$ \\
Composite concept~\cite{Habibian:ICMR14} & $5.97$\\
Weak concept~\cite{Wu:CVPR14} & $3.48$\\
Annotated concept~\cite{SRI:NIST13}& $6.50$\\
Our EventNet concept & $\mathbf{8.86}$\\
\hline
\end{tabular}
\end{center}
\caption{Comparisons between our ECR with other state-of-the-art concept based video representation methods built on visual content. All results are obtained in the task of zero-shot event retrieval on TRECVID MED 2013 test set. }
\label{Tab:ComWithStateArt}
\end{table}

\begin{figure} [htp]
\centering
\includegraphics[width=1\columnwidth]{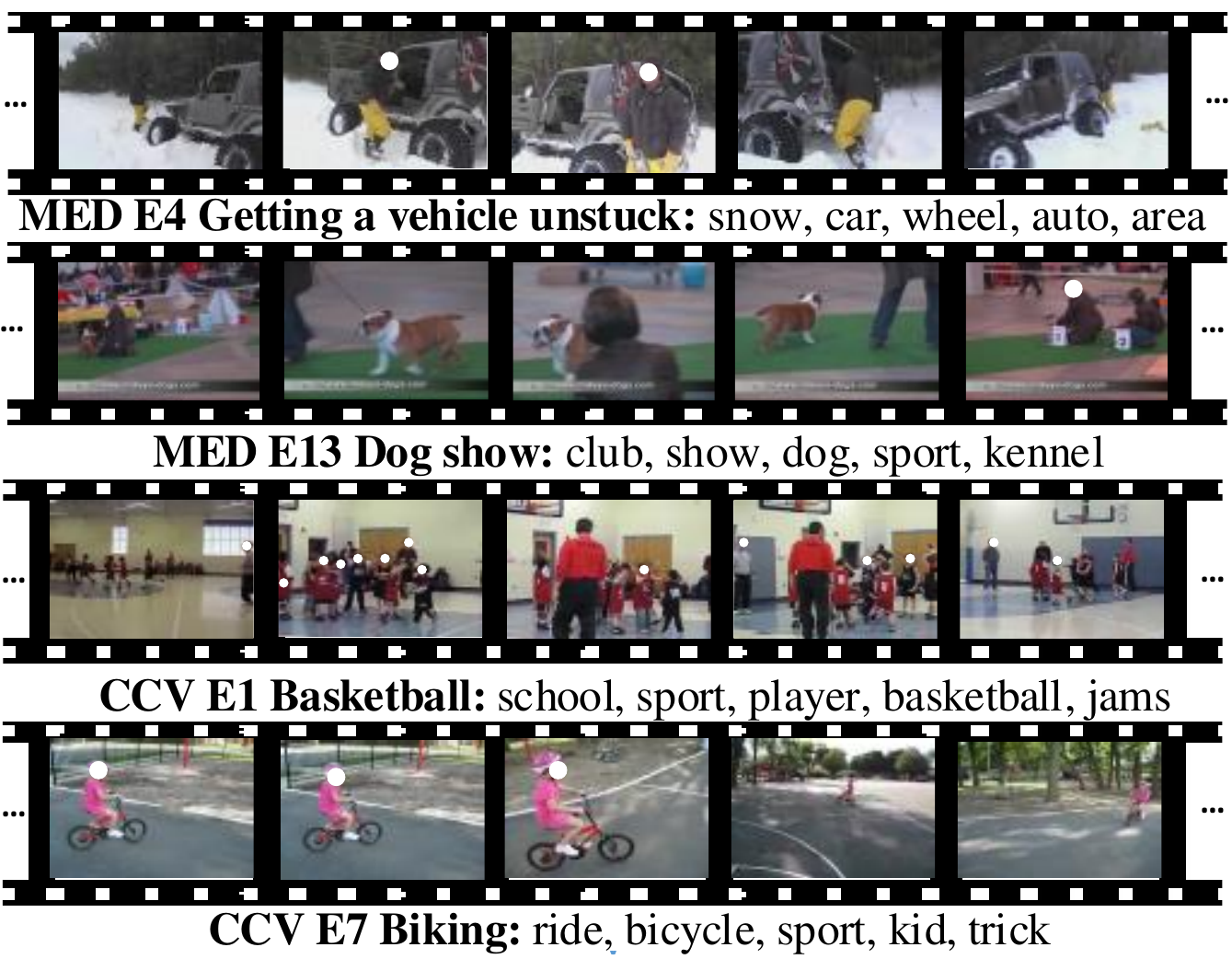}
\caption{Event video recounting results: each row shows evenly subsampled frames of a video and the top $5$ concepts detected in the video. }
\label{fig_recounting}
\end{figure}

\subsection{Task II: Semantic Recounting in Videos}
Given a video, semantic recounting aims to annotate the video with the semantic concepts detected in the video. Since we have the concept based representation generated for the videos using the concept classifiers described earlier, we can directly use it to produce the recounting. Specifically, we rank the $4,490$ event specific concept scores on a given video in descending order and then choose the top ranked ones as the most salient concepts occurring in the video. Figure~\ref{fig_recounting} shows the recounting results for some example videos from TRECVID MED and CCV datasets. As seen, the concepts generated by our method precisely reveal the semantics presented in the videos.

It is worth noting that EventNet ontology also provides great benefits for developing a real time semantic recounting system which requires high efficiency and accuracy. Compared to other concept libraries that use generic concepts, EventNet allows selected execution of a small set of concepts specific to an event. Given a video to be recounted, it first predicts the most relevant events and then applies only concepts specific to these events. This unique two-step approach is able to greatly improve the efficiency and accuracy of multimedia event recounting since only a small number of event-specific concept classifiers need to be fired after event detection.


\subsection{Task III: Effects of EventNet Structure for Concept Matching}\label{Sec:StructureEval}
As discussed in Section~\ref{sec:semanticmatching}, due to the limitations of text based similarity matching, the matching result of an event query might not be relevant. In this case, the EventNet structure can help the users find relevant events and their associated concepts from EventNet concept library. Here we first carry out quantitative empirical study to verify this. Specifically, for each event query, we manually specify two suitable categories from the top $19$ nodes of EventNet tree structure, and then match events under these categories based on semantic similarity. We compare the results obtained by matching all events in EventNet (i.e., without leveraging EventNet structure) with the results obtained by the method we described above (i.e., with leveraging EventNet structure). For each query, we apply each method to select $15$ concepts from EventNet library and then use them to perform zero-shot event retrieval.

\begin{table}[htp]
\begin{center}
\begin{tabular}{l||c|c}
Method (mAP \%) & MED & CCV\\
\hline\hline
Without Leveraging EventNet Structure  & $8.86$ & $35.58$\\
With Leveraging EventNet Structure & $\mathbf{8.99}$ & $\mathbf{36.07}$\\
\hline
\end{tabular}
\end{center}
\caption{Comparison of zero-shot event retrieval using the concepts matched without leveraging EventNet structure (top row) and with leveraging EventNet structure (bottom row).}
\label{Tab:ComStructure}
\end{table}

Table~\ref{Tab:ComStructure} shows the performance comparison between the two methods. From the results, we can see that event retrieval performance can be improved if we apply the concepts matched with the help of EventNet structure, which proves the usefulness of EventNet structure for the task of concept matching.


\begin{figure}[t]
\begin{center}
\includegraphics[width=1\linewidth]{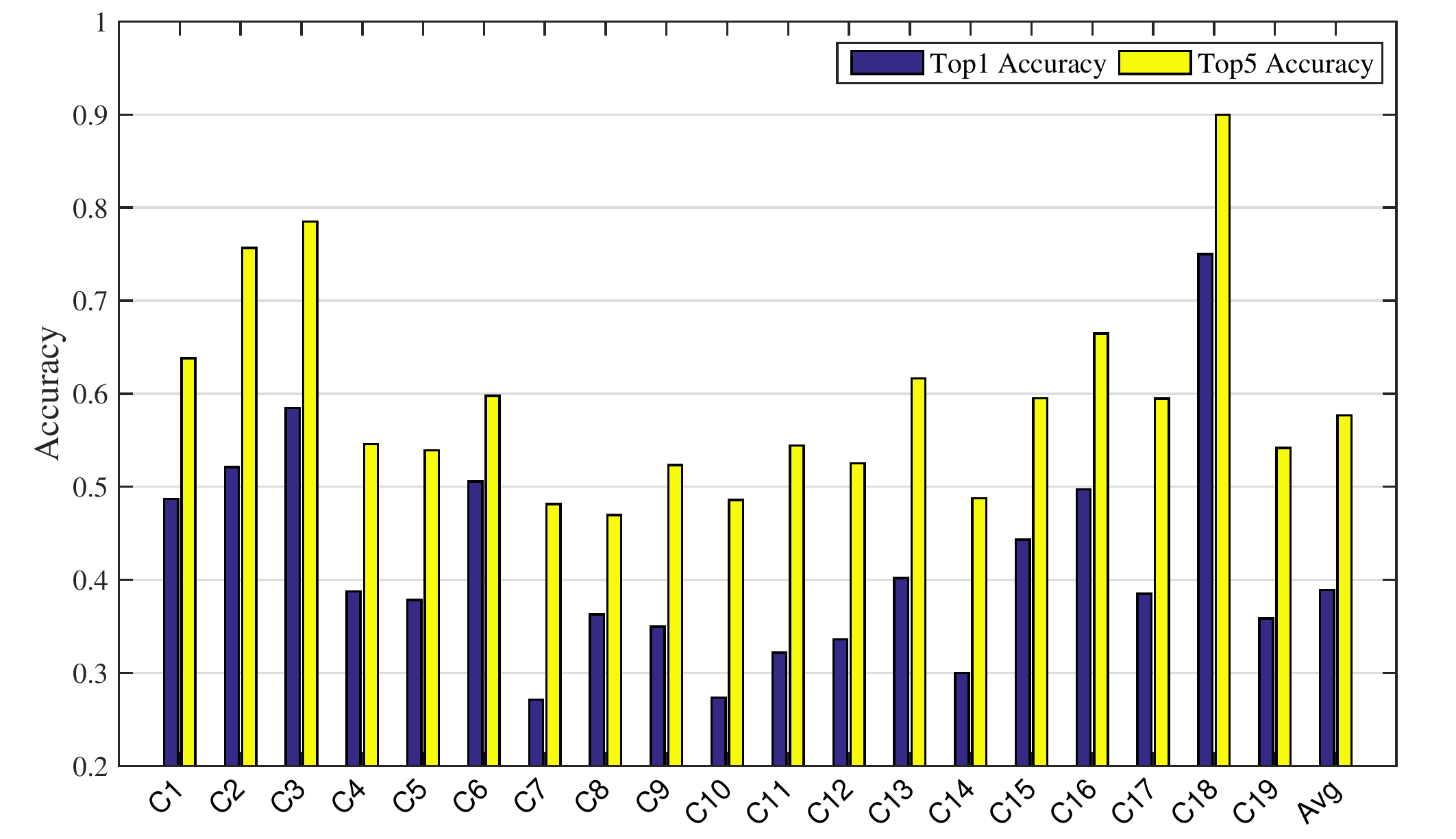}
\caption{Top-$1$ and top-$5$ event classification accuracies over $19$ high-level event categories of EventNet structure, in which the average top-1 and top-5 accuracy are $38.91\%$ and $57.67\%$.}
\label{Fig:Top15}
\end{center}
\end{figure}

\begin{figure}[htp]
\begin{center}
\includegraphics[width=1\linewidth]{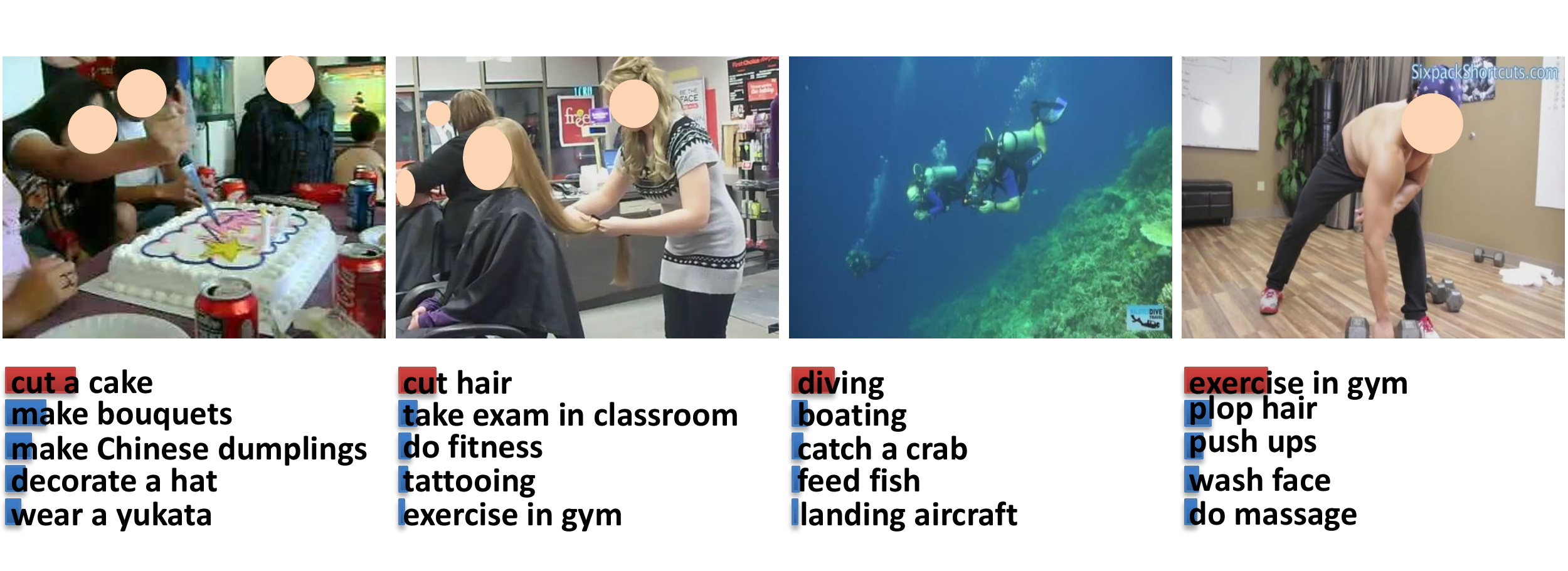}
\caption{Event detection results of some sample videos. The $5$ events with the highest detection scores are shown in the descending order. The bar length indicates the score of each event. Event with the red bar is the ground truth.}
\label{Fig:PredResult}
\end{center}
\end{figure}

\subsection{Task IV: Multi-Class Event Classification}
The $95,321$ videos over $500$ event categories in EventNet can also be seen as a benchmark video dataset to study large scale event detection. To facilitate direct comparison, we provide standard data partitions and some baseline results over these partitions. It is worth noting that one important purpose of designing EventNet video dataset is to use it as a testbed for large scale event detection models such as deep convolutional Neutral Network. Therefore, in the following, we summarize a baseline implementation using the state-of-the-art CNN models like what was done in~\cite{Russakovsky:arXiv14}.

\textbf{Data Split}. Recall that each of the $500$ events in EventNet has around $190$ videos. In our experiment, we split videos and adopt $70\%$ videos as training set,
$10\%$ videos as validation set, and $20\%$ videos as test set. In all, there are roughly $70K$ training videos ($2.8$ million frames), $10K$ validation videos ($0.4$ million frames) and $20K$ test videos ($0.8$ million frames).

\textbf{Deep Learning Model}. We adopt the same network architecture and learning setting of the CNN model described in Section~\ref{Sec:CNN} as our multi-class event classification model. In the training process, for each event, we treat the frames sampled from training videos of an event as positive training samples and feed them into CNN model for model training. It takes seven days to finish $450K$ iterations of training. In the test stage, to produce predictions for a test video, we take the average of the frame-level probabilities over sampled frames in a video and use it as the video-level prediction result.

\textbf{Evaluation Metric}. Regarding evaluation metric, we adopt the most popular top-$1$ and top-$5$ accuracy commonly used in large scale visual recognition, where the top-1 (top-5) accuracy is the fraction of the test videos for which the correct label is among the top $1$ ($5$) labels predicted to be most probable by the model.

\textbf{Results}. We report the multi-class classification performance by $19$ high-level categories of events in the top layer of EventNet ontology. To achieve this, we collect all events under each of the $19$ high-level categories in EventNet (e.g., $68$ events under ``home and garden''), calculate the accuracy of each event and then calculate their mean value over the events within this high-level category. As seen in Figure~\ref{Fig:Top15}, most of the high-level categories show impressive classification performance. To illustrate the results, we choose four event video frames and show their top-$5$ prediction results in Figure~\ref{Fig:PredResult}. 


\section{Conclusion}
We have introduced EventNet, a large scale structured event-driven concept library for representing complex event in video. The library contains $500$ events mined from WikiHow and $4,490$ event-specific concepts discovered from YouTube video tags, for which large margin classifiers are trained with deep learning feature over 95,321 YouTube videos. The events and concepts are further organized into a tree structure based on the WikiHow ontology. Extensive experiments on two benchmark event datasets have shown major superior performance of the proposed concept library over zero-shot event retrieval task. We also show that the tree structure of EventNet helps matching the event queries to semantically relevant concepts. For future work, we will continue to expand EventNet by continuously discovering more events from WikiHow, YouTube and other knowledge resources. We will also pursue a tree structured event modeling that incorporates the hierarchical relations of events in EventNet.

\section{Acknowledgement}
Supported by the Intelligence Advanced Research Projects Activity (IARPA) via Department of Interior National Business Center contract number D11PC20071. The U.S. Government is authorized to reproduce and distribute reprints for Governmental purposes notwithstanding any copyright annotation thereon. Disclaimer: The views and conclusions contained herein are those of the authors and should not be interpreted as necessarily representing the official policies or endorsements, either expressed or implied, of IARPA, DOI-NBC, or the U.S. Government.



\balancecolumns
\end{document}